\documentclass[10pt,twocolumn,letterpaper]{article}

\usepackage[pagenumbers]{cvpr} 

\usepackage[dvipsnames]{xcolor}

\usepackage{times}
\usepackage{epsfig}
\usepackage{graphicx}
\usepackage{amsmath}
\usepackage{amssymb}
\usepackage{lipsum}
\usepackage{multirow}

\definecolor{cvprblue}{rgb}{0.21,0.49,0.74}
\usepackage[pagebackref,breaklinks,colorlinks,citecolor=cvprblue]{hyperref}

\title{TaoAvatar: Real-Time Lifelike Full-Body Talking Avatars for \\ Augmented Reality via 3D Gaussian Splatting}

\author{Jianchuan Chen\footnotemark[1],\quad Jingchuan Hu\footnotemark[1],\quad Gaige Wang\footnotemark[1],\quad Zhonghua Jiang\\
Tiansong Zhou,\quad Zhiwen Chen$^\ddag$,\quad Chengfei Lv$^\dag$\\
Alibaba Group, Hangzhou, China\\
}

\begin{document}
\twocolumn[
\maketitle
\vspace{-4.0em}
\begin{center}
\includegraphics[width=1\linewidth]{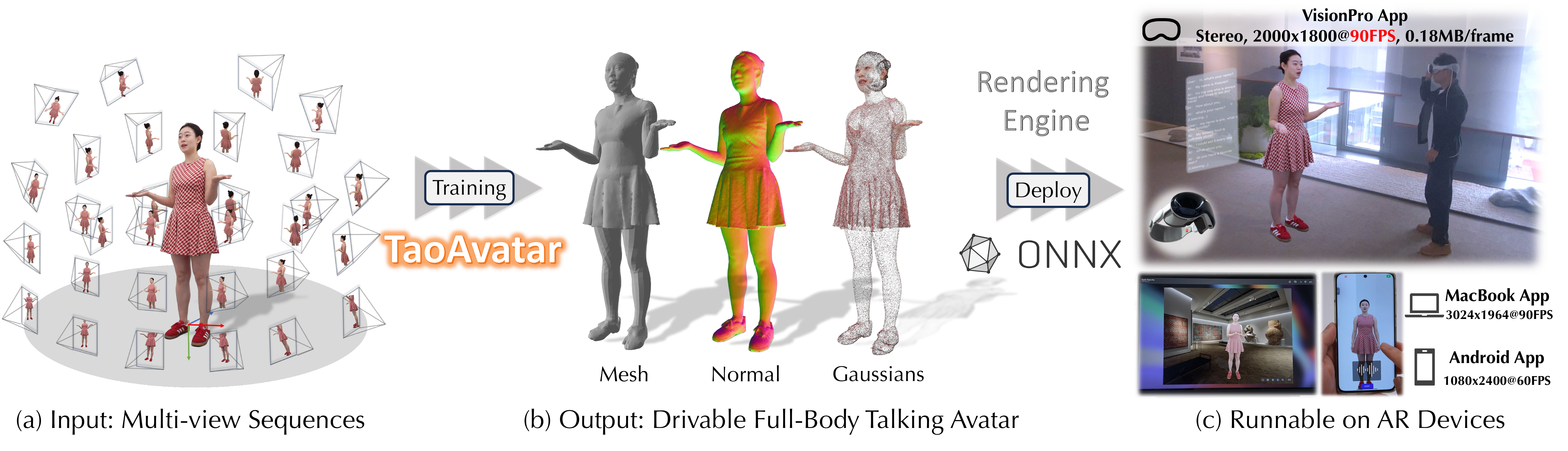}
\end{center} \vspace{-1.8em}
\captionof{figure}{TaoAvatar generates photorealistic, topology-consistent 3D full-body avatars from multi-view sequences. It provides high-quality, real-time rendering with low storage requirements, compatible across various mobile and AR devices like the Apple Vision Pro.}
\label{fig:teaser}


\bigbreak]

\renewcommand{\thefootnote}{\fnsymbol{footnote}}
\footnotetext[1]{Equal contribution. $^\ddag$Project Leader and $^\dag$Corresponding author. Project: \url{https://PixelAI-Team.github.io/TaoAvatar}}


\begin{abstract}
Realistic 3D full-body talking avatars hold great potential in AR, with applications ranging from e-commerce live streaming to holographic communication. Despite advances in 3D Gaussian Splatting (3DGS) for lifelike avatar creation, existing methods struggle with fine-grained control of facial expressions and body movements in full-body talking tasks.
Additionally, they often lack sufficient details and cannot run in real-time on mobile devices.
We present TaoAvatar, a high-fidelity, lightweight, 3DGS-based full-body talking avatar driven by various signals.
Our approach starts by creating a personalized clothed human parametric template that binds Gaussians to represent appearances.
We then pre-train a StyleUnet-based network to handle complex pose-dependent non-rigid deformation, which can capture high-frequency appearance details but is too resource-intensive for mobile devices.
To overcome this, we "bake" the non-rigid deformations into a lightweight MLP-based network using a distillation technique and develop blend shapes to compensate for details.
Extensive experiments show that TaoAvatar achieves state-of-the-art rendering quality while running in real-time across various devices, maintaining 90 FPS on high-definition stereo devices such as the Apple Vision Pro.

\end{abstract}

\vspace{-1.0em}
\section{Introduction}

Creating lifelike 3D human avatars is a dynamic and rapidly advancing research area in computer vision and graphics, essential for applications in AR/VR, 3D entertainment, and virtual communication.
Despite decades of research, achieving realistic and immersive avatars on mobile devices—particularly AR—remains challenging due to computational limitations. Current industry solutions, such as MetaHuman, often rely on high-precision scans and extensive manual effort by artists for 3D modeling, rigging, and motion capture.

In recent years, academic researchers have adopted parametric models~\cite{smpl,smplx} for human representation, which are constructed from massive body scans and can capture a performer's shape, expression, and pose from images or videos.
These methods~\cite {SMPLD,VideoAvatar,multi_garment_net,SMPLX_Lite} extend the naked model~\cite{smpl,smplx} by adding offsets to represent tailored clothing, but are still struggling with complex geometries and high-frequency details, such as loose skirts and fine hair, due to limitations of topology and texture resolution. 
With the rise of neural radiance fields (NeRF)~\cite{nerf}, many researchers tend to define implicit neural human representations in canonical space, animated by backward linear blend skinning~\cite{zhao2022humannerf,peng2021animatable} of these parametric models~\cite{smpl,smplx}.
While these methods can handle arbitrary topology and deliver higher rendering quality, they are highly time-consuming due to the large amount of sampling required during volume rendering.
Although many efforts~\cite{instantavatar,fast_snarf} have significantly improved rendering efficiency by introducing iNGP~\cite{instantngp}, the animated results still fail to escape the Uncanny Valley effect when driven.
Recently, 3D Gaussian Splatting (3DGS)~\cite{3dgs}, an explicit and efficient point-based representation, has gained tremendous attention from researchers for its ability to deliver both high-quality rendering and real-time performance.
Compared to implicit representations~\cite{nerf,neus2}, the explicit point-based representation is more compatible with being integrated with the mesh-based parametric models~\cite{smpl,smplx}, which can be directly driven by forward linear blend skinning.
Many recent methods~\cite{gaussianavatar,3dgs-avatar,splatting_avatar,animatable_gaussian}, combining 3DGS and the parametric models~\cite{smpl,smplx}, have made significant strides toward realizing lifelike 3D human avatars.
However, achieving real-time rendering on mobile devices remains challenging, especially for AR glasses, which require high resolution, stereo rendering, and seamless interaction.

We introduce \textbf{TaoAvatar}, a high-fidelity, lightweight, 3DGS-based full-body talking avatar designed for running in real-time on Augmented Reality (AR) devices.
Unlike previous approaches~\cite{SNARF,animatable_gaussian,meshavatar}, which develop implicit parameterized models from scratch, we construct a personalized, clothed parameterized template that preserves the facial expressions and gesture controls inherent to the native SMPLX~\cite{smplx}.
At the same time, we bind the Gaussians to the triangles as texture to create a hybrid avatar representation.
Inspired by~\cite{animatable_gaussian}, we pre-train a large StyleUnet-based teacher network to learn pose-dependent dynamic deformation maps of Gaussians in 2D space by front and back orthogonal projection.
Although the teacher network is powerful enough to capture high-frequency appearance details in various poses, its large number of parameters makes it difficult to run in real-time on AR devices.
To achieve high-performance rendering, we bake the dynamic deformation of Gaussians into the non-rigid deformation field of the mesh using a distillation technique, employing a compact MLP-based student network.
Meanwhile, we propose two lightweight, learnable blend shapes to compensate for the non-rigid deformation of the Gaussians.
After fine-tuning, our student model achieves high-performance rendering without compromising quality.

To evaluate our approach, we propose a multi-view dataset named \textbf{TalkBody4D}, focusing on prevalent full-body talking scenarios encountered in daily life, which includes rich facial expressions and gestures with synchronous audios different from existing full-body motion datasets~\cite{humanrf, DNA-Rendering, AvatarReX}.
Experimental results demonstrate that our method effectively captures high-frequency appearance details of human avatars while maintaining a lightweight architecture capable of rendering stereo images at 2K resolution on common mobile and AR devices, including the Apple Vision Pro.
The resulting full-body avatar is highly expressive, enabling users to animate it with facial expressions, hand gestures, and body poses, thereby highlighting the potential applications of our method.
Our contributions can be summarized as follows:
\begin{itemize}
\item  We introduce \textbf{TaoAvatar}, a novel teacher-student framework which yields topologically consistent, well-aligned geometry, further creating a high-fidelity, lightweight, 3DGS-based drivable full-body talking avatar.

\item We propose two innovative strategies including non-rigid deformation \textbf{baking} and two lightweight blend shapes \textbf{compensations} to ensure high-quality rendering and efficient performance on mobile and AR devices, achieving \textbf{2K} resolution rendering at \textbf{90 FPS} on high-definition stereo devices like the Apple Vision Pro.
\item We contribute \textbf{TalkBody4D}, a multi-view dataset to be released, designed for full-body talking scenarios, featuring diverse facial expressions and gestures with synchronous audios.
Extensive experiments demonstrate that our approach outperforms other state-of-the-art methods in both quality and performance.
\end{itemize}

\section{Related work}

\noindent \textbf{3D Avatar Representation.}
Traditional computer graphics techniques~\cite{stoll2010video, vlasic2009dynamic, de2007marker, marc2021} use 3D scanning or multi-view stereo (MVS) to create full-body meshes, which are then combined with a 3D skeleton to produce driveable templates. However, this approach is both expensive and inefficient.
As parametric body models, SMPL~\cite{smpl} and SMPLX~\cite{smplx} are widely used for their versatility but lack detail in simulating loose clothing. 
Some works~\cite{SMPLD, VideoAvatar, SMPLX_Lite} have attempted to add clothing offsets to SMPLX, effectively handling tight clothing but struggling with loose garments.
Due to the limited expressiveness of traditional meshes, researchers have combined SMPL with advanced rendering techniques.
Some methods~\cite{AnimNeRF, HumanNeRF, SelfRecon, NeuralActor} use NeRF \cite{nerf} to implicitly express human bodies, mapping points from the observation space backward to canonical space.
Although high-quality, volume rendering is slow and poses challenges for real-time applications.
In contrast, 3DGS \cite{3dgs} uses 3D Gaussian distributions for scenes, offering real-time rendering \cite{luiten2023dynamic, Wu_2024_CVPR}, fast training, and the ability to handle complex materials such as hair and translucency. 
Some methods \cite{3dgs-avatar, animatable_gaussian, splatting_avatar, GoMAvatar, GAvatar} combining SMPLX~\cite{smplx} and 3DGS~\cite{3dgs}, like GaussianAvatar \cite{gaussianavatar}, use explicit GS point clouds and forward skinning to efficiently simulate the deformations, offering good rendering quality and speed.
This representation presents potential solutions for achieving both real-time performance and computational efficiency.

\noindent \textbf{Dynamic Nonrigid Deformation Modeling.}
Various methods have been developed to simulate non-rigid human deformation realistically.
The MLP-based methods \cite{GauHuman, HUGS, GART, HumanGaussianSplatting}, like 3DGS-Avatar \cite{3dgs-avatar}, are noted for lightweight flexibility but struggle with handling novel poses.
Physical simulation methods \cite{physavatar, Dressing_Avatars} produce physics-based animations but require unified topology, and complex modeling, and are computationally expensive, which negatively impacts real-time performance.
Additionally, some methods \cite{HOOD, ContourCraft} use GNNs to learn non-rigid deformations in a data-driven manner.
These methods are faster than traditional cloth simulations.
Recently, dynamic texture methods are emerging, these methods \cite{NeuralActor, KwonGeneralizable, uvgaussian} model complex details in the 2D texture space, while AnimatableGS~\cite{animatable_gaussian} and MeshAvatar \cite{meshavatar} enhance modeling with front and back projection strategies which avoid texture unwarping.

\noindent \textbf{3D Drivable Talking Avatars.}
Research on drivable digital humans is rapidly advancing for various applications.
In 3D Talking Heads, 
methods~\cite{gaussian_avatars, Gaussian_Head_Avatar, Gaussian_Codec_Avatars, PSAvatar, FlashAvatar, HeadGaS} excel in capturing and driving facial expressions, enhancing the realism of virtual characters.
For drivable bodies, these methods~\cite{gaussianavatar, INSTA_Avatar, animatable_gaussian, SNARF, GoMAvatar} achieve precise restoration of body movements by learning pose-dependent deformation fields.
Furthermore, some methods \cite{XAvatar, Dressing_Avatars} not only support detailed body modeling and animation but also enable head expressions and lip-syncing.
Despite enhancing realism, running these methods in real-time on mobile or AR devices is challenging due to their high computational demands, exceeding current edge platform capabilities.
Optimizing computational efficiency without compromising high-quality output is a key research focus.


\section{Method}

\begin{figure*}[ht]
    \centering
    \includegraphics[width=1.0\linewidth]{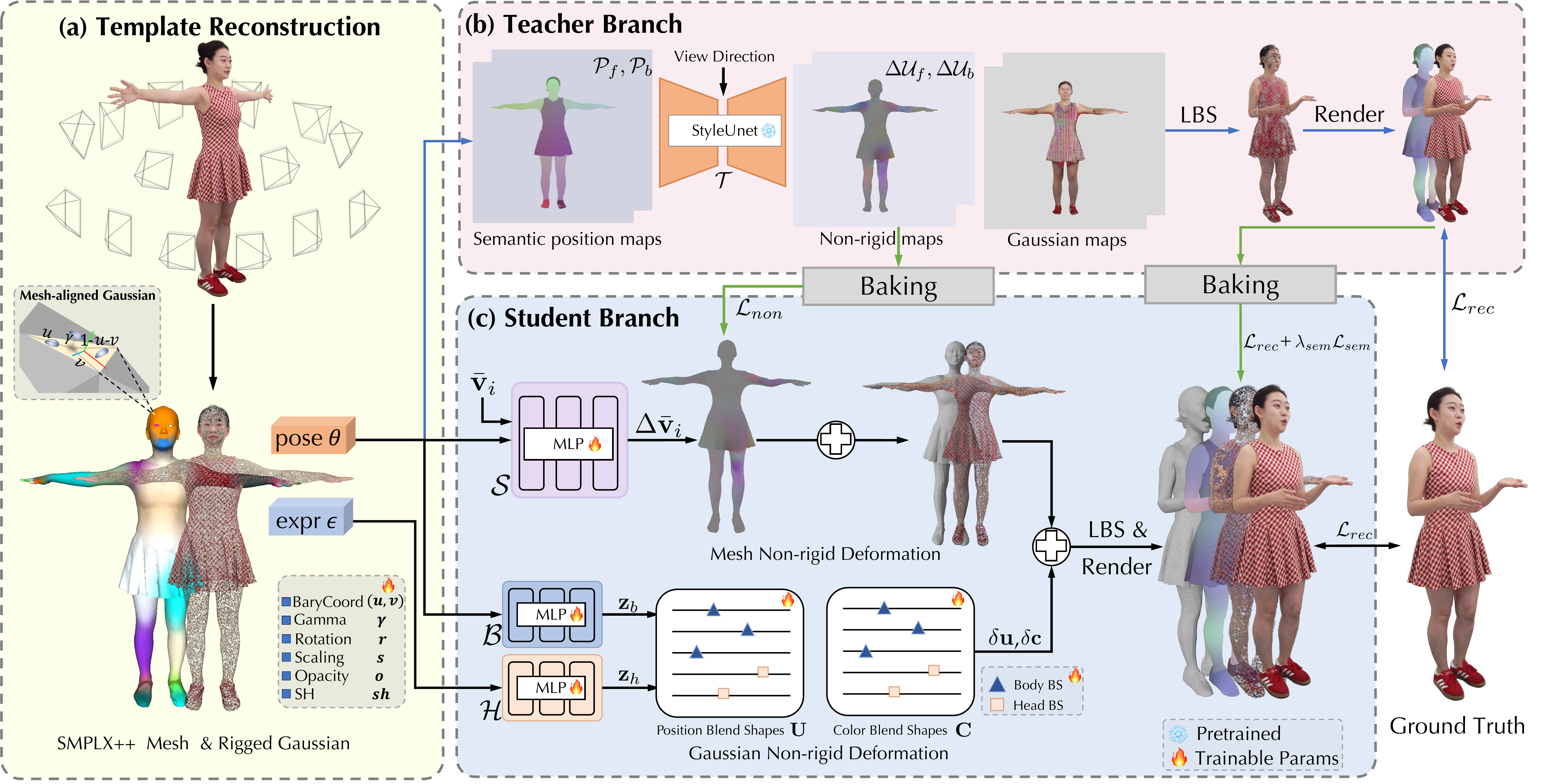}
    \vspace{-1.5em}
    \caption{\textbf{Illustration of our Method.}
Our pipeline begins by reconstructing (a) a cloth-extended SMPLX mesh with aligned Gaussian textures. To address complex dynamic non-rigid deformations, (b) we employ a teacher StyleUnet to learn pose-dependent non-rigid maps, which are then baked into a lightweight student MLP to infer non-rigid deformations of our template mesh. For high-fidelity rendering, (c) we introduce learnable Gaussian blend shapes to enhance appearance details.}
    \label{fig:method}
\end{figure*}

Leveraging multi-view RGB videos of a performer alongside corresponding SMPLX registrations for each frame,
we aim to develop a high-fidelity, lightweight full-body talking avatar capable of high-resolution and high-performance rendering on mobile devices.
To achieve this, we first create a hybrid clothed parametric representation by integrating 3D Gaussian Splatting (3DGS)~\cite{3dgs} and SMPLX~\cite{smplx}, which is more expressive for simulating loose clothing and hair, while maintaining high-quality rendering (\cref{smplx_plus_plus}).
Next, we propose a teacher-student framework to reconstruct high-frequency, pose-dependent dynamic details as illustrated in~\cref{fig:method}.
By leveraging non-rigid deformation baking and incorporating two lightweight blend shape compensations, our method achieves a tradeoff between high quality and high performance (\cref{gs_maps}).

\subsection{Hybrid Clothed Parametric Representations}

\noindent \textbf{The Clothed Extension of SMPLX.}
\label{smplx_plus_plus}
To extend SMPLX to complete clothed human geometry with clothes, hair, etc., we choose a frame close to T-pose as a reference, which provides more visible details and less sticky geometry.
First, we utilize NeuS2~\cite{neus2} to reconstruct the geometry of this reference frame and parse out non-body components (e.g., clothes) from it using~\cite{4ddress}.
However, the body's skeleton cannot directly drive these components.
We estimate the shape and pose of the SMPLX~\cite{smpl} in the current frame and use robust skinning transfer~\cite{robust_skin_weights_transfer} to propagate the body’s skinning weights to these non-body components.
Finally, we apply inverse skinning to transform these components back to the standard T-pose, resulting in a comprehensive personalized parametric model, referred to as \textbf{SMPLX++}.

\noindent \textbf{Binding Gaussians on Mesh as Texture.}
\label{gs_texture}
After obtaining the complete geometric model, we bind 3D Gaussians to the mesh's triangles as textures, enabling them to move synchronously with the mesh.
Inspired by these hybrid representations~\cite{gaussian_avatars,splatting_avatar}, we define the attributes of the Gaussians in a local coordinate system relative to the triangles.
Specifically, for each triangle, we randomly initialize $k$ Gaussians, each of which maintains the local attributes $\{f, (u, v), \gamma, \mathbf{r}, \mathbf{s}, o, \mathbf{sh}\}$, where $f$ is the index of the parent triangle, $(u, v)$ is the barycentric coordinates, $\gamma$ represents the translation along the triangle's normal direction, $\mathbf{r}$ and $\mathbf{s}$ denote rotation and scaling in the local space, $\mathbf{o}$ indicates opacity, and $\mathbf{sh}$ consists of spherical harmonic coefficients.
Additionally, we define the normal of each Gaussian as $\mathbf{n}_g = [1, 0, 0]$ within the local space, ensuring consistency with the mesh's normal for subsequent transformations.
To deform each Gaussian from the local to the world space, we construct a transformation based on its parent triangle $F_f(\mathbf{v}_1, \mathbf{v}_2, \mathbf{v}_3)$, where $\mathbf{v}_1$, $\mathbf{v}_2$, $\mathbf{v}_3$ are the vertices of the triangle $F_f$ in the world space:
\begin{equation}
\begin{aligned}
\mathbf{p} &= u\cdot \mathbf{v}_1 + v \cdot \mathbf{v}_2 + \left(1-u-v\right)\cdot \mathbf{v}_3,\\
\mathbf{R} &= [\mathbf{n},\mathbf{q},\mathbf{n}\times \mathbf{q}], \mathbf{q} = \frac{\left(\mathbf{v}_2+\mathbf{v}_3 \right)/2 - \mathbf{v}_1}{\left\|\left(\mathbf{v}_2+\mathbf{v}_3\right)/2 - \mathbf{v}_1\right\|},\\
e &= \left(\left\|\mathbf{v}_1-\mathbf{v}_2 \right\| + \left\|\mathbf{v}_2-\mathbf{v}_3 \right\|  + \left\|\mathbf{v}_1-\mathbf{v}_3 \right\|\right)/3,
\end{aligned}
\end{equation}
where $\mathbf{n} =\frac{\left(\mathbf{v}_1-\mathbf{v}_2\right) \times\left(\mathbf{v}_3-\mathbf{v}_1\right)}{\left\|\mathbf{v}_1-\mathbf{v}_2\right\| \cdot\left\|\mathbf{v}_3-\mathbf{v}_1\right\|}$ is the normal of the triangle, $\mathbf{p}$ denotes the surface point on the parent triangle, and $e$ denotes the average edge length.
\begin{equation}
\mathbf{u}_w = \mathbf{p} + \gamma\mathbf{R}\mathbf{n},\quad
\mathbf{r}_w = \mathbf{R}\mathbf{r},\quad
\mathbf{s}_w = e \cdot \mathbf{s},
\label{eq:local_to_world_transformation}
\end{equation}
here, $\mathbf{u}_w$, $\mathbf{r}_w$, $\mathbf{s}_w$ represent the position, rotation, and scaling in the world spaces after transformation, respectively. Furthermore, $\mathbf{c}_w = SH\left(\mathbf{sh}, \mathbf{R}^{-1} \mathbf{d} \right) $ denotes the color along the view direction $\mathbf{d}$ in the world space.

\subsection{Dynamic Mesh-based Gaussian Reconstruction}

\noindent \textbf{Learning Dynamic Non-Rigid Gaussian Deformation Maps.}
\label{gs_maps}
Although SMPLX++ can be directly driven by the skeleton, linear blend skinning is insufficient for performing dynamic non-rigid deformation such as clothing folds and swaying motions.
Like dynamic Gaussian avatars~\cite{animatable_gaussian,uvgaussian}, we utilize a large StyleUnet~\cite{styleavatar} as the teacher network to capture complex pose-dependent dynamic non-rigid deformation of Gaussians in 2D texture space.
Following~\cite{animatable_gaussian}, we rasterize the T-pose mesh, which is colored using posed coordinates and segmentation colors, to obtain the front and back position maps $\{\mathcal{P}_f, \mathcal{P}_b\}$. These position maps are then fed into the StyleUnet network along with view directions to generate the non-rigid deformation maps $\{\Delta\mathcal{U}_f, \Delta\mathcal{U}_b\}$ and other residual Gaussian attribute maps.
We add these residuals to the Gaussian's local attributes and transform them into the world space according to Eq.~\eqref{eq:local_to_world_transformation}.
To train the StyleUnet-based teacher network $\mathcal{T}$, we adopt the losses including L1, D-SSIM~\cite{3dgs}, and perceptual loss~\cite{lpips}.
Additionally, we introduce a normal loss $\mathcal{L}_{nor}$.
Consequently, the total reconstruction loss $\mathcal{L}_{rec}$ is defined as:
\begin{equation}
    \mathcal{L}_{rec} = \mathcal{L}_{1} + \lambda_{ssim}\mathcal{L}_{ssim} + \lambda_{lpips}\mathcal{L}_{lpips} + \lambda_{nor}\mathcal{L}_{nor},
\end{equation}
where $\lambda_{*}$ is the loss weights. The normal loss is defined as $\mathcal{L}_{nor} = \left\| N_t - N_g \right\|$, where $N_t$ is the rendered normal image by the teacher network, and the ground truth normal image $N_g$ can be obtained from the method \cite{neus2}.

\noindent \textbf{Baking Non-Rigid Gaussian Deformation into Mesh Deformation.}
\label{baking_nonrigid}
Although the StyleUnet-based network $\mathcal{T}$ can achieve good results, it struggles to run in real-time on mobile devices due to its large number of parameters, especially for AR devices, which require high resolution, stereo rendering, and high performance. 
Inspired by knowledge distillation, we bake the Gaussian non-rigid deformation of the teacher network into the mesh non-rigid deformation field, which is a compact MLP-based student network $\mathcal{S}$.
\begin{equation}
    \Delta\bar{\mathbf{v}}_i = \mathcal{S}\left(\bar{\mathbf{v}}_i, \mathbf{\theta}, \mathbf{z}_{t} \right),
\end{equation}
where $\bar{\mathbf{v}}_i$ is the $i$-th vertex coordinate in the canonical space, $\mathbf{\theta}$ is the pose parameter, and $\mathbf{z}_{t}$ is a learnable embedding for each frame to compensate for inaccurate pose estimation.
To train the student network $\mathcal{S}$ with the capability of pose-dependent non-rigid deformation, we directly supervise its output using the Gaussian non-rigid deformation maps $\{\Delta\mathcal{U}_f, \Delta\mathcal{U}_b\}$ of the teacher network $\mathcal{T}$.
Specifically, we render the mesh non-rigid deformations $\left\{\Delta\mathbf{v}_i\right\}_{i=1}^N$ to front and back deformation maps $\{\Delta\mathcal{V}_f,\Delta\mathcal{V}_b\}$ in the canonical space, and then calculate the differences:
\begin{equation}
    \mathcal{L}_{non} = \left\| \Delta\mathcal{V}_f - \Delta\mathcal{U}_f \right\| + \left\| \Delta\mathcal{V}_b - \Delta\mathcal{U}_b \right\|.
\end{equation}
Furthermore, to mitigate the intersection between clothing and the body, we introduce a semantic loss $\mathcal{L}_{sem}$.
For each vertex, we assign a semantic label $\mathbf{e}_i = \mathbf{c}_i + \sin\left({\tau\bar{\mathbf{v}}_i}\right)$ that integrates the candidate coordinates $\bar{\mathbf{v}}_i$ and the artificial segmentation color $\mathbf{c}_i$, and $\tau$ is a scale factor. The semantic label for each Gaussian can be obtained by interpolating the semantic label of the three vertices of its parent triangle.
Subsequently, we render the Gaussian semantic map $\mathcal{E}_t$ and the mesh semantic map $\mathcal{E}_s$ in the world space separately and compute the semantic loss $\mathcal{L}_{sem}$:
\begin{equation}
    \mathcal{L}_{sem} = \left\| \mathcal{E}_s - \mathcal{E}_t \right\|.
\end{equation}

In addition, we refine the local attributes by computing the reconstruction loss $\mathcal{L}_{rec}$, which takes the rendered image of the teacher network as ground truth.
The total loss $\mathcal{L}_{bak}$ during baking is:
\begin{equation}
    \mathcal{L}_{bak} = \mathcal{L}_{rec} + \lambda_{non} \mathcal{L}_{non} + \lambda_{sem} \mathcal{L}_{sem}.
\end{equation}

\noindent \textbf{Compensating Gaussian Deformation with Lightweight Blend Shapes.} 
\label{bs_nonrigid}
Inspired by pose corrective blend shapes in SMPL~\cite{smpl}, we build learnable blend shapes for position and color for each Gaussian.
The position blend shapes, $\mathbf{U} \in \mathbb{R}^{3\times n}$, primarily compensate for Gaussian positional adjustments (e.g., hair fluttering, clothing folds) across different poses. Meanwhile, the color blend shapes, $\mathbf{C} \in \mathbb{R}^{3\times n}$, address appearance changes, such as shadow variations caused by self-obscuration.
To obtain the corresponding driving coefficients, we employ a mapping network $\mathcal{H}$ for the head using the expression parameters $\epsilon$ as input and a mapping network $\mathcal{B}$ for the body which inputs the body pose parameters $\mathbf{\theta}$, to get the coefficients $\mathbf{z}_{h} \in \mathbb{R}^{n_h} $ for the head and $\mathbf{z}_{b} \in \mathbb{R}^{n_b}$ for the body, respectively, enabling independent and fine-grained control over head and body deformations.
The compensation of position and color for each Gaussian can be expressed as:
\begin{equation}
\begin{aligned}
\delta \mathbf{u} &= BS\left(\mathbf{z}_{h} \oplus \mathbf{z}_{b}; \mathbf{U} \right), \\
\delta \mathbf{c} &= BS\left(\mathbf{z}_{h} \oplus \mathbf{z}_{b}; \mathbf{C} \right),
\end{aligned}
\end{equation}
where $\oplus$ is a concatenation operation, resulting in the size $n = n_h + n_b$. The function $BS(\cdot,\cdot)$ represents the blend shapes operation mentioned in SMPL~\cite{smpl}.
We add these residuals to local attributes:
\begin{equation}
\begin{aligned}
\mathbf{u}_w &= \mathbf{p} + \mathbf{R}\left(\gamma \mathbf{n} + \delta \mathbf{u}\right), \\
\mathbf{c}_w &= SH\left(\mathbf{R}^{-1} \mathbf{d} \right) + \delta \mathbf{c}.
\end{aligned}
\label{eq:local_to_world_transformation_with_residuals}
\end{equation}
For the fine-tuning stage on training data, we freeze the non-rigid deformation field network $\mathcal{S}$, optimize the two mapping networks $\mathcal{H}$ and $\mathcal{B}$, and two blend shapes $\mathbf{U}$ and $\mathbf{C}$.
The loss during fine-tuning is the same as that pre-training the teacher network.

After baking mesh non-rigid deformation and applying Gaussian non-rigid compensation, the performance of our student model is significantly enhanced compared to the teacher model, while still maintaining high quality.

\section{Experiments}

\begin{figure*}[ht]
    \centering
    \includegraphics[width=1.0\linewidth]{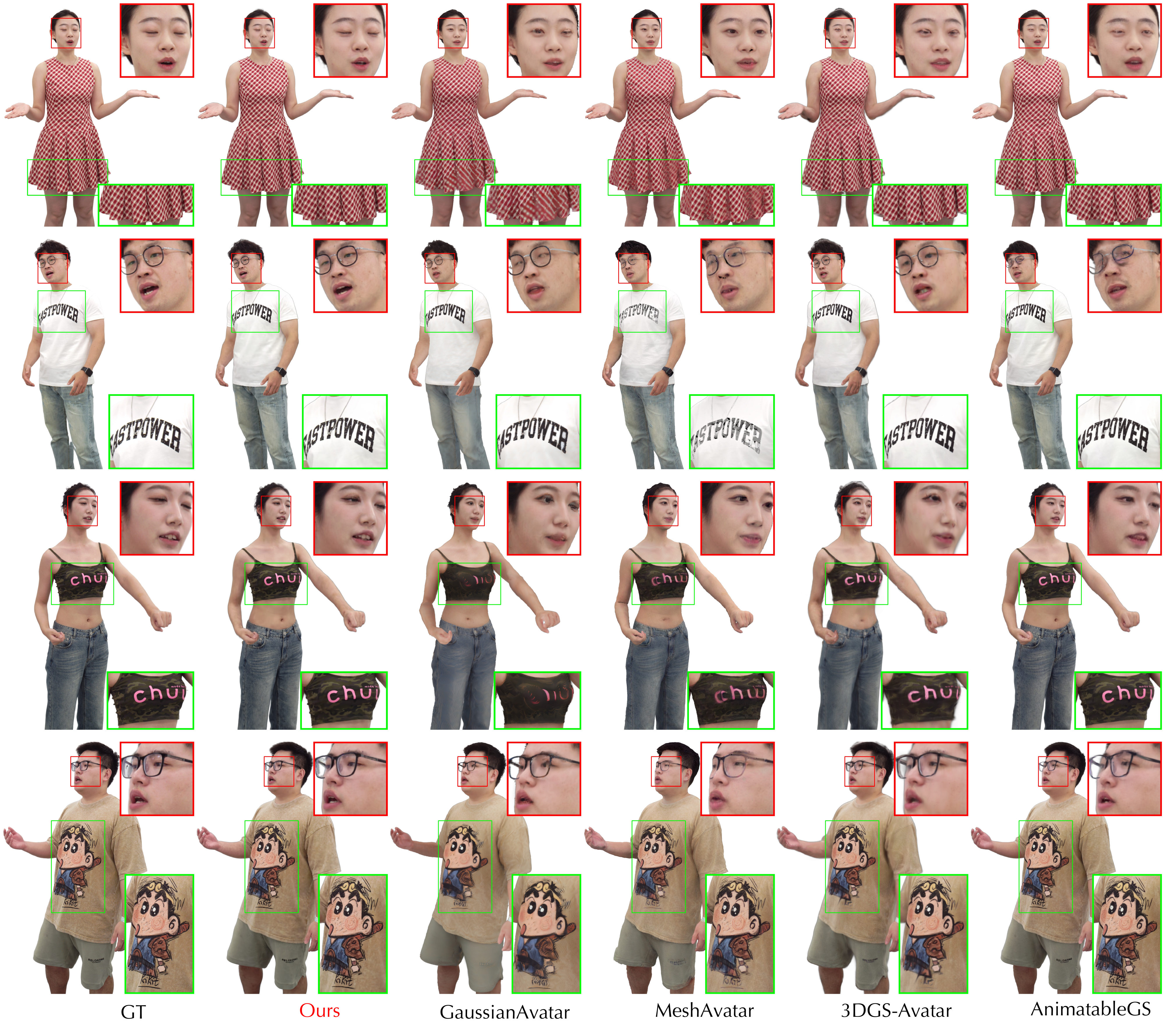}
    \vspace{-2.5em}
    \caption{\textbf{Qualitative comparisons on full-body talking tasks.} Our method outperforms state-of-the-art methods by producing clearer clothing dynamics and enhanced facial details.}
    \label{fig:novel_view_on_talkingbody}
    \vspace{-1.0em}
\end{figure*}

\subsection{Experimental Settings}

\noindent \textbf{Datasets.}
\label{dataset description}
To evaluate the performance on the full-body talking task, we introduce a new dataset, \textbf{TalkBody4D}, designed for common full-body talking scenarios in everyday life, primarily characterized by rich mouth movements and diverse gestures with synchronous audios. 
The dataset comprises $4$ distinct identities, each wearing $2$ different outfits.
Each talking sequence consists of rough $6$k frames with $4$K image resolution, $60$ views with $48$ full-body, and $12$ face-region views.
To evaluate performance on complex motions and expressions, we select $4$ sequences (04, 05, 06, 08) from the \textbf{ActorHQ}~\cite{humanrf} dataset, supplement with $2$ dancing and $2$ exaggerated expression sequences.
The resolution of our training and testing images is $1500\times 2000$ for all experiments.
SMPLX registrations for these sequences can be obtained by first initializing SMPLX parameters using existing tools~\cite{easymocap,vhap,gaussian_avatars}, followed by fine-tuning the poses with a photometric loss to improve geometry alignment.

\noindent \textbf{Implementation Details.}
The clothed parametric template has roughly $22k$ vertices and $45k$ faces in total, which contains additional $23k$ faces for clothes, hair, and shoes, compared to the naive SMPLX.
We randomly initialize about $220k$ Gaussians in total to bind with the template as texture, for each triangle containing roughly $4$ to $6$, and optimize these Gaussians for $10k$ iterations using the multi-view images of the reference frame.
We pre-train our StyleUnet-based\cite{animatable_gaussian} teacher network for $600k$ iterations, and set the loss weights $\lambda_{ssim}=0.2,\lambda_{lpips}=0.01,\lambda_{nor}=0.02$ and batch size is $1$ during the training process.
Then we bake the teacher network into a small $5$-layer MLP-based student network, which set $\lambda_{non}=0.1,\lambda_{sem}=1.0$ and optimize $30k$ iterations.
Finally, we finetune the mapping networks and blend shapes for Gaussian non-rigid compensation on training data $100k$ iterations at the batch size of $4$.
The size of two blend shapes for position and color is $n=28$, in which $n_h=8$ for the head and $n_b=20$ for the body.

\subsection{Comparison}

We conduct an extensive comparison between our method and recent state-of-the-art full-body avatar methods include GaussianAvatar~\cite{gaussianavatar}, 3DGS-Avatar~\cite{3dgs-avatar}, MeshAvatar~\cite{meshavatar}, and AnimatableGS~\cite{animatable_gaussian}.
All of these methods learn pose-dependent non-rigid deformations in the canonical space and then utilize the skeleton-based rigid transformations of SMPL~\cite{smpl} or SMPLX~\cite{smplx} into observation space to model the dynamic human. We uniformly use SMPLX~\cite{smplx} across all these methods to ensure a fair comparison.

\noindent \textbf{Comparison on Full-body Talking.}
We conduct experiments on typical full-body talking scenarios encountered in daily life, which include a variety of mouth movements and gestures.
We provide the quantitative comparison on full-body talking as shown in~\cref{tab:novel_view_and_novel_pose_on_talking_body}.
We also show the qualitative comparison in~\cref{fig:novel_view_on_talkingbody}, and our method can generate more realistic and detailed results, especially in the face region.
3DGS-Avatar~\cite{3dgs-avatar} directly learns two MLP-based networks to handle the pose-dependent non-rigid deformation of Gaussians 
but produces very blurry results due to the low-frequency bias of MLPs. GaussianAvatar~\cite{gaussianavatar} uses a 2D CNN network by encoding the position map in SMPLX's UV space to generate the non-rigid deformation maps of Gaussians. However, it is restricted by the mini-clothed topology of SMPLX, struggling to handle loose skirts.
MeshAvatar~\cite{meshavatar} chooses mesh as a human representation, which is insufficient to express the geometry of details, such as hair, and glasses.
AnimatableGS~\cite{animatable_gaussian} can generate high-quality dynamic appearance and complex non-rigid deformation.
However, it is restricted by the implicit template of learning from scratch, which makes it difficult to handle fine-grained expressions, like blinking and flashing teeth as shown in the face region in~\cref{fig:novel_view_on_talkingbody}.
Therefore we adopt the SMPLX++ model as the template whose face and hands retain the native SMPLX prior.
However, due to the amount parameters of the teacher StyleUnet~\cite{styleavatar}, it achieves poor performance and is hard to run on a mobile device in real time.
Through baking non-rigid deformation, we can achieve $150+$ FPS at the resolution of $1500\times 2000$ on Nvidia RTX4090 using only a small MLP-based mesh non-rigid deformation network and two lightweight learnable blend shapes as shown in~\cref{tab:novel_view_and_novel_pose_on_talking_body}.

\begin{table*}[ht]
	\centering
	\scalebox{0.90}{
		\begin{tabular}{c|ccc|ccc|c}
			\toprule

                &\multicolumn{3}{c|}{Novel View}
			&\multicolumn{3}{c|}{Novel Gestures and Expression}
                &\multicolumn{1}{c}{Speed}
   
			\\
			\midrule
			
			Model & PSNR$\uparrow$ & SSIM$\uparrow$ & LPIPS$\downarrow$   & PSNR$\uparrow$ & SSIM$\uparrow$ & LPIPS$\downarrow$ & FPS \\
			\midrule
			
			GaussianAvatar\cite{gaussianavatar} & 26.58 (23.57) & .9313 (.8159) & .10577 (.25242) & 25.99 (23.15)  & .9232 (.8092) & .12265 (.26207) & 54 \\
			
			3DGS-Avatar\cite{3dgs-avatar} & 28.91 (23.95) & .9411 (.8303) & .07984 (.20450) & 26.46 (23.32) & .9157 (.8184) & .11804 (.21632) & 55 \\
			MeshAvatar\cite{meshavatar} & 28.53 (24.55)  & .9360 (.8083) & .09470 (.25572) & 27.08 (23.58) & .9229 (.7965) & .10783 (.24947) & 22 \\
                AnimatableGS\cite{animatable_gaussian}  & 32.50 (26.42) & .9599 (.8587) & .06695 (.19535) & 28.05 (23.68) & .9328 (.8142) & .09191 (.22673) &  16 \\
   
			Ours (Teacher) & \underline{33.45} (\underline{27.01}) & \underline{.9649} (\underline{.8741}) & \textbf{.04986} (\underline{.15613}) & \underline{28.28} (\underline{24.28}) & \underline{.9336} (\underline{.8291}) & \textbf{.07385} (\underline{.18176}) & 16   \\
   
			Ours (Student) & \textbf{33.81} (\textbf{27.80}) & \textbf{.9689} (\textbf{.8975}) & \underline{.06437} (\textbf{.14218}) & \textbf{28.38} (\textbf{24.99}) & \textbf{.9389} (\textbf{.8525}) & \underline{.08874} (\textbf{.13364}) &  \textbf{156} \\
			
			\bottomrule
	\end{tabular}}
	\caption{{\bf Quantitative comparisons on full-body talking task}. The results inside the parentheses are evaluated for the face area, and the inference speed is evaluated on Nvidia RTX4090 when rendering images at a resolution of 1500 × 2000.}
    \vspace{-1em}
	\label{tab:novel_view_and_novel_pose_on_talking_body}
\end{table*}
\begin{figure*}[ht]
    \centering
    \includegraphics[width=1.0\linewidth]{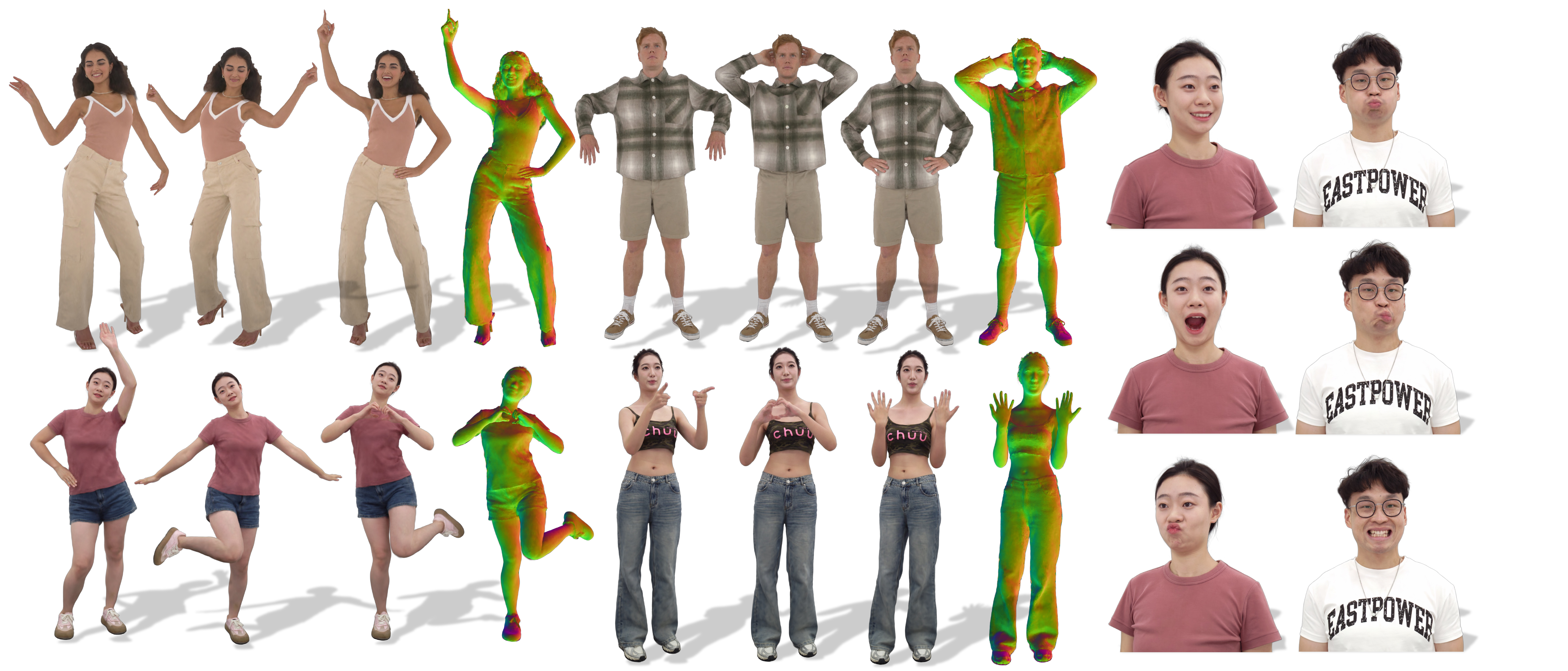}
    \vspace{-2.0em}
    \caption{\textbf{Results in challenging scenarios.} Our method can obtain high-quality reconstruction even for challenging pose and expressions.}
    \label{fig:novel_view_on_motion_and_expression}
\end{figure*}
\begin{table}[ht]
	\centering
	\scalebox{0.80}{
		\begin{tabular}{c|ccc}
			\toprule
						
			Model & PSNR$\uparrow$ & SSIM$\uparrow$ & LPIPS$\downarrow$ \\
			\midrule
			
			GaussianAvatar\cite{gaussianavatar} & 25.94 (24.33) & .9294 (.8251) & .10478 (.24179) \\
			
			3DGS-Avatar\cite{3dgs-avatar} & 30.04 (25.08)  & .9403 (.8458) & .08471 (.18044) \\
			
			MeshAvatar\cite{meshavatar} & 28.51 (24.94) & .9334 (.8100) & .08846 (.23517) \\
			AnimatableGS\cite{animatable_gaussian}  & 31.81 (26.79) & .9493 (.8608) &  .07586 (.19521)  \\
   
			Ours (Teacher) & \textbf{32.80} (\textbf{27.40}) & \underline{.9533} (\underline{.8768}) & \textbf{.05581} (\underline{.14996}) \\
   
			Ours (Student) & \underline{32.72} (\underline{27.35}) & \textbf{.9579} (\textbf{.8836}) & \underline{.07326} (\textbf{.13914}) \\
			
			\bottomrule
	\end{tabular}}
	\caption{{\bf Quantitative comparisons about complex motions and expressions reconstruction}. The results inside the parentheses are evaluated for face regions with exaggerated expressions.}
    \vspace{-1em}
	\label{tab:novel_view_on_motion_and_expression}
\end{table}

\noindent \textbf{Comparison on Complex Motion and Expression.}
Besides full-body talking scenes, our method is also capable of reconstructing complex motions as well as exaggerated expressions as shown in~\cref{fig:novel_view_on_motion_and_expression}. 
We provide quantitative comparisons in ~\cref{tab:novel_view_on_motion_and_expression}, our student network achieves competitive results compared to the teacher network, outperforming other methods.
Especially for exaggerated expressions, our method significantly outperforms others in reconstructing the face region as shown in ~\cref{tab:novel_view_on_motion_and_expression}.
Moreover, our method can obtain high-quality normals as shown in ~\cref{fig:novel_view_on_motion_and_expression}, which can be used for image-based relighting.

\begin{figure}[ht]
    \centering
    \includegraphics[width=1.0\linewidth]{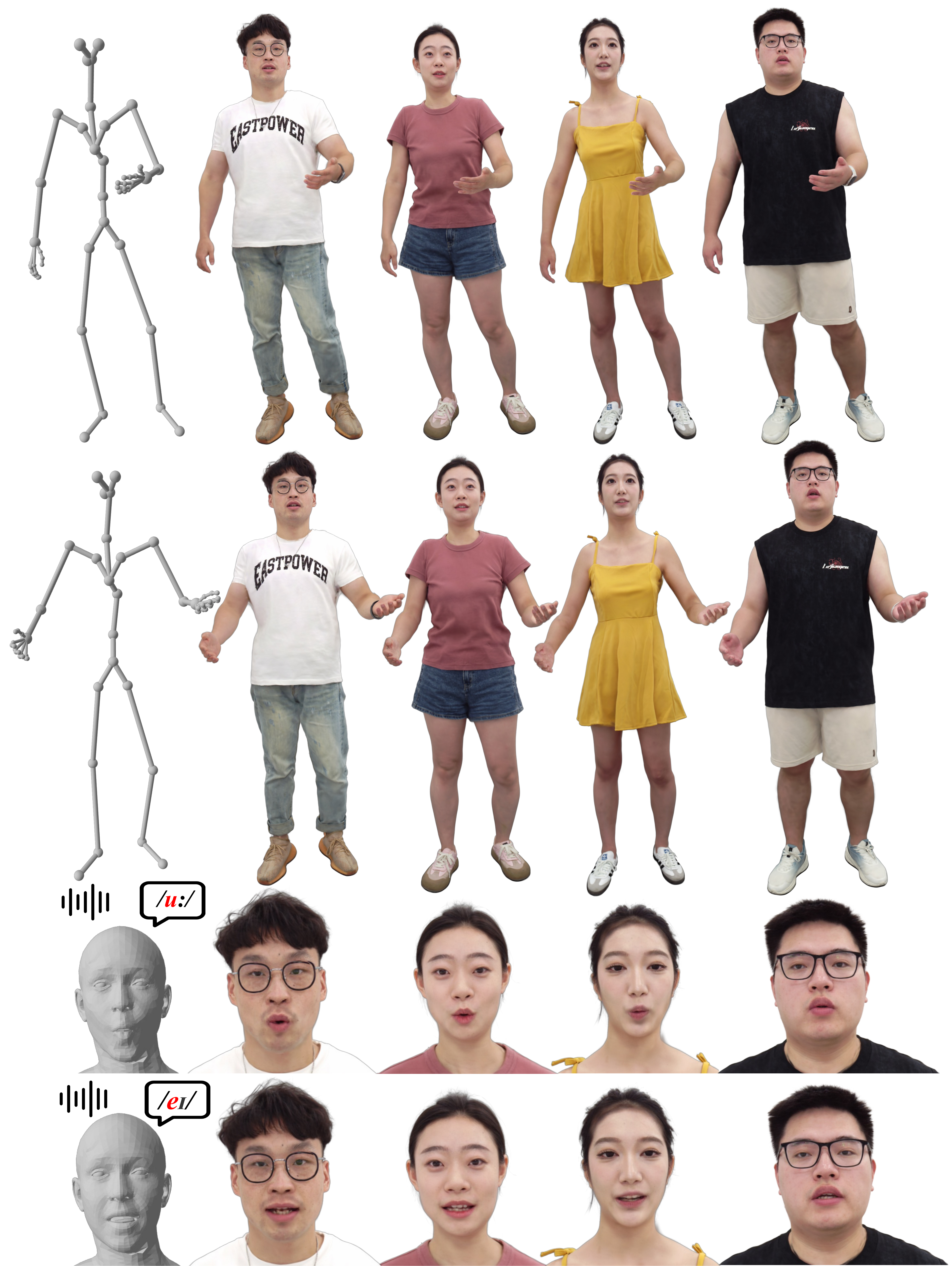}
    \vspace{-1.5em}
    \caption{\textbf{Novel pose and expression animation.} TaoAvatar can be driven by the same skeleton and expression parameters vividly.}
    \label{fig:novel_pose}
    \vspace{-1.0em}
\end{figure}

\noindent \textbf{Comparison on Novel Expression and Novel Gesture.} 
Benefiting from our parametric human model SMPLX++, our full-body avatars can be expression-driven and gesture-driven.
We provide quantitative comparisons of self-driven on full-body talking scenes as shown in~\cref{tab:novel_view_and_novel_pose_on_talking_body}.
As shown in~\cref{fig:novel_pose}, we provide visualizations of different characters driven by the same skeleton.
In addition, our model can accept expression parameter inputs generated from audio by UniTalker~\cite{unitalker} as shown in~\cref{fig:novel_pose}.
More demos are provided in the supplementary materials.

\subsection{Ablation Study}

In this subsection, we discuss the key contributions of our method. Additional quantitative results and experiments are available in the supplementary materials.

\begin{figure}[ht]
    \centering
    \includegraphics[width=1.0\linewidth]{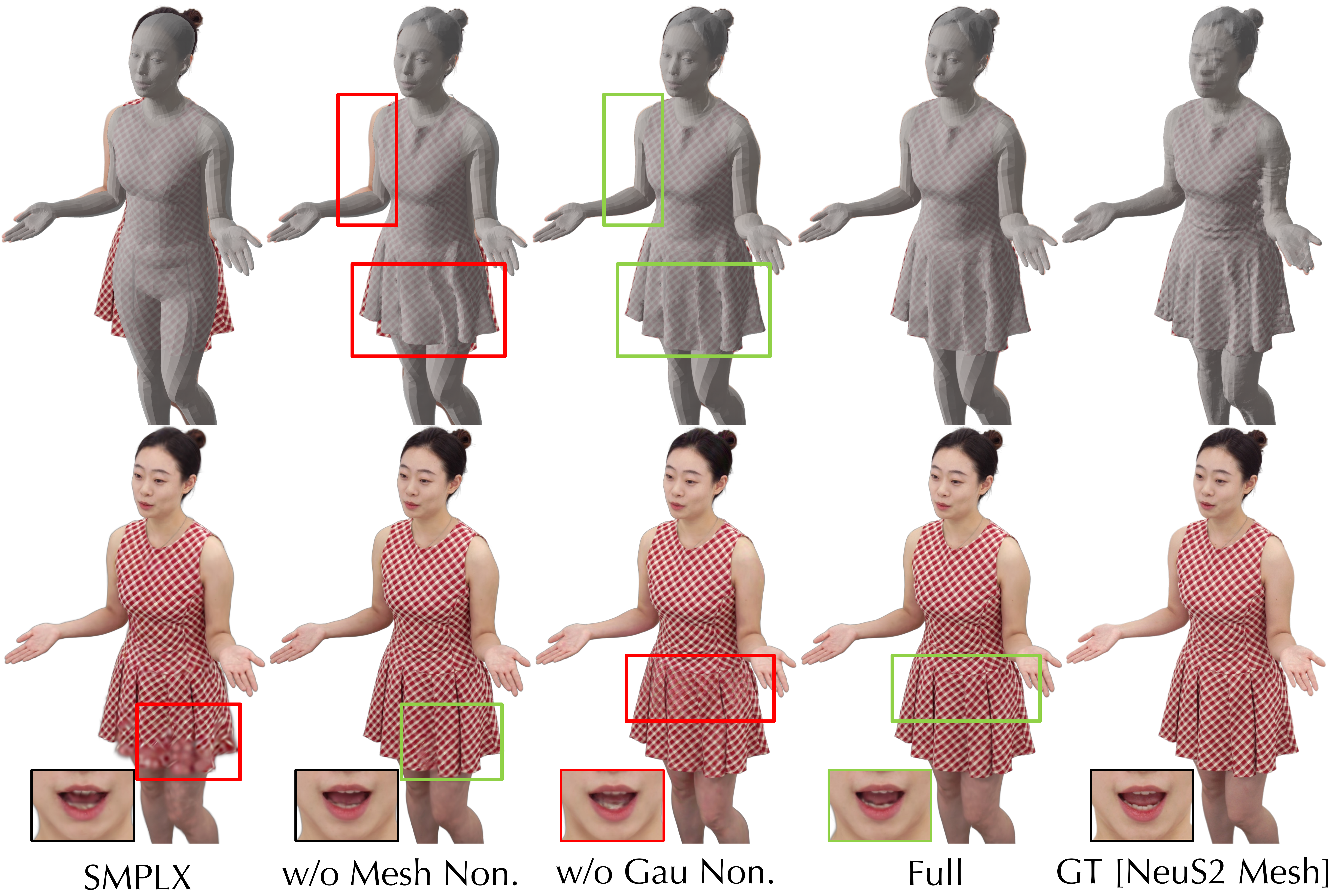}
    \vspace{-1.5em}
    \caption{Ablation Study. \textcolor{red}{Red} and \textcolor{green}{Green} boxes show artifacts and their improved counterparts.}
    \label{fig:ablation_study}
    \vspace{-0.0em}
\end{figure}

\noindent \textbf{The Impact of Template.}
The choice of template has a crucial impact on our approach. If we directly use SMPLX as our geometric template, as shown in~\cref{tab:ablation_study} (w SMPLX), leading to poor results with numerous artifacts, especially for loose skirts as seen in~\cref{fig:ablation_study}.
After extending the geometry of SMPLX to a complete template with clothes, hair, and shoes, this quality is significantly improved as shown in~\cref{tab:ablation_study} (w/o Mesh Non.), which allows our method to reconstruct more challenging clothes.

\noindent \textbf{The Impact of Non-rigid Deformation.}
Despite having a complete geometry, the rigid transformations of the skeleton are not able to properly represent non-rigid transformations such as skirt swinging and hair flying. We concurrently introduce mesh non-rigid deformation and Gaussian non-rigid deformation in our method. The geometry without mesh non-rigid deformation is unable to attach to the surface of the character as shown in~\cref{fig:ablation_study} (w/o Mesh Non.).
The Gaussian non-rigid compensation also plays a crucial role in quantitative metrics. Without the blend shapes for Gaussian non-rigid compensation, the results are prone to produce artifices due to imprecise geometry and variable appearance as shown in~\cref{fig:ablation_study} (w/o Gau Non.).

\noindent \textbf{The Impact of Baking Teacher.}
Compared to directly training a small MLP-based network to handle non-rigid deformations as shown in~\cref{tab:ablation_study} (w/o Teacher), our proposed multi-stage training strategy is more efficient.
With direct supervision of the non-rigid deformation of the teacher network, it is easier for the student network to decouple the geometry deformation and appearance variations.

\section{Application}
\begin{table}[ht]
	\centering
	\scalebox{0.85}{
		\begin{tabular}{c|ccc|cc}
			\toprule
						
			Model & PSNR$\uparrow$ & SSIM$\uparrow$ & LPIPS$\downarrow$ & P2S$\downarrow$ & Chamfer$\downarrow$ \\
			\midrule
			
			w SMPLX & 28.47 & .9476 & .07899 & .7690 & .9995 \\
			
			w/o Mesh Non. & 32.10 & .9734 & .03814 & .4877 & .5007 \\
   
			w/o Gau Non.  & 31.16 & .9686 & .03932 & .2968 & .3068 \\
   
			w/o Teacher & 32.67 & .9751 & .03769 & .5236 & .5359 \\
   
			Full & \textbf{33.29} & \textbf{.9772}  & \textbf{.03464} & \textbf{.2953} & \textbf{.3052} \\
			
			\bottomrule
	\end{tabular}}
	\vspace{-0.5em}
	\caption{{\bf Ablation Study}. We use the mesh reconstructed by NeuS2~\cite{neus2} as a pseudo-truth for geometry evaluation.}
    \vspace{-1.0em}
	\label{tab:ablation_study}
\end{table}
TaoAvatar offers a lightweight, comprehensive representation for 3D talking body scenarios, easily deployable on various mobile devices using existing tools~\cite{MNN} as demonstrated in~\cref{fig:teaser}. It seamlessly integrates with AI models to enable real-time dialogue interactions. we deployed a 3D digital human agent on the Apple Vision Pro, which interacts with users through an ASR-LLM-TTS pipeline~\cite{Paraformer, Qwen2, VITS2, Walle}. Facial expressions and gestures are dynamically controlled by an Audio2BS model~\cite{unitalker}, allowing the agent to respond naturally with synchronized speech, expressions, and movements. A live demonstration is available in our supplementary materials.


\section{Discussion}

\noindent \textbf{Conclusion.}
In this work, we present TaoAvatar, a lightweight, lifelike full-body talking avatar solution. We demonstrate how the teacher-student framework captures high-definition facial and body details while ensuring real-time performance on AR devices. TaoAvatar can be driven by diverse signals, including facial expressions, hand gestures, and body poses. Through both quantitative and qualitative evaluations, we showcase the advantages of our approach. Additionally, we validate its practical potential with a real-world application on the Apple Vision Pro.

\noindent \textbf{Limitations and Future Works.}
TaoAvatar encounters challenges in modeling flexible clothing deformation under exaggerated body poses, which are out-of-distribution of training data. A possible solution is to integrate GNN simulators~\cite{HOOD,ContourCraft} handling larger hemlines, which are compatible with our approach.

\noindent \textbf{Potential Social Impact.}
TaoAvatar can synthesize lifelike talking digital humans within an augmented reality environment, generating fabricated 3D content or 2D videos. Therefore, responsible use of this technology is essential.


\clearpage

{
    \small
    \bibliographystyle{ieeenat_fullname}
    \bibliography{references}
}

\begin{appendix}

\twocolumn[
\vspace{-3.5em}
\begin{center}
\includegraphics[width=1\linewidth]{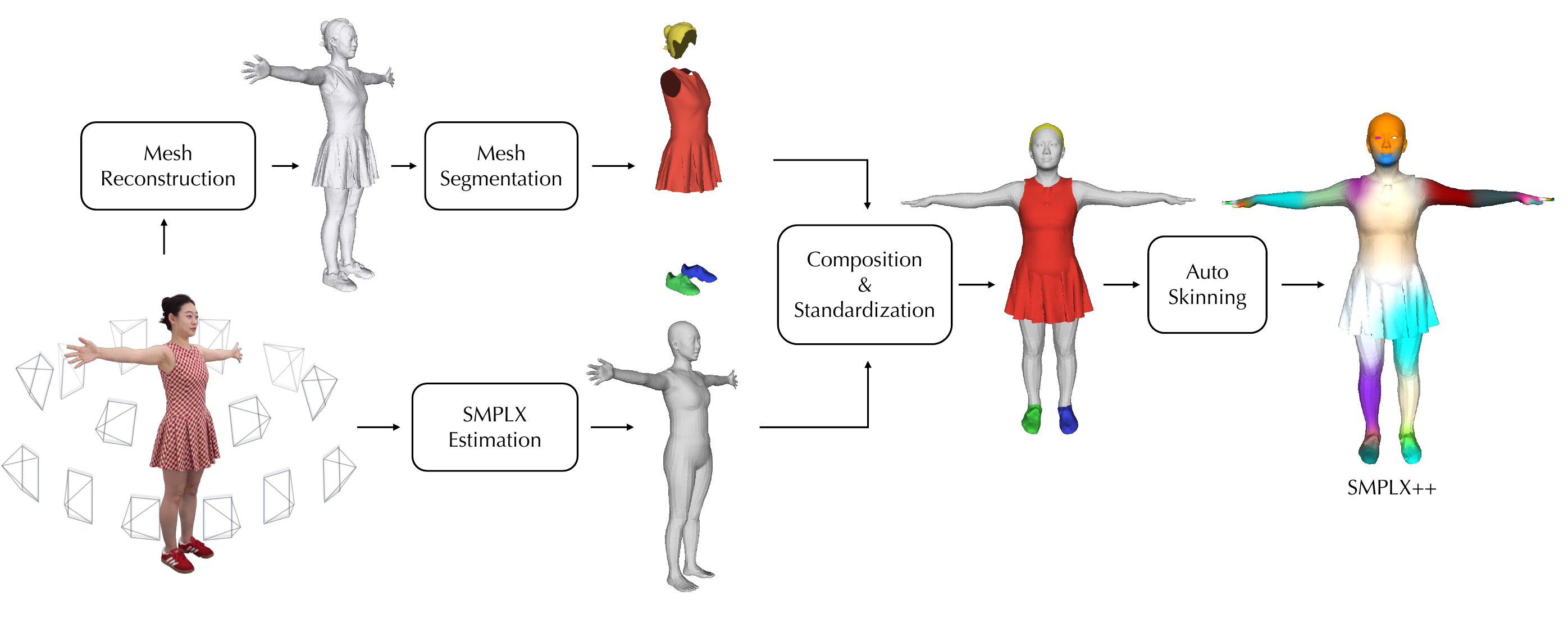}
\end{center} \vspace{-2.0em}
\captionof{figure}{\textbf{The Pipeline of the Template SMPLX++ Reconstruction.}}
\label{fig:template_reconstruction}
\bigbreak]


\section{Implementation Details}
\noindent \textbf{Template Reconstruction.}
The clothed template SMPLX++ plays a vital role in our approach.
As shown in~\cref{fig:template_reconstruction}, we create a pipeline to obtain the personalized parametric template SMPLX++ from multi-view images.
We choose a frame close to T-pose as a reference, providing more visible details and less sticky geometry and making obtaining accurate SMPLX parameters easier.
First, we reconstruct the complete geometry from the multi-view images using NeuS2~\cite{neus2}.
Then we segment and simplify the non-body components such as skirt, shoes, and hair, according to the method proposed in 4D-Dress~\cite{4ddress}.
However these components are not under the standard T-pose space, we estimate the SMPLX parameters for the reference frame using existing tools~\cite{easymocap,vhap}, and transform them back to T-pose space according to the inverse rigid transformation.
Specifically, these skinning weights for non-body parts can be automatically generated by Robust Skinning Transfer~\cite{robust_skin_weights_transfer}.
Finally, we combine the naked SMPLX with segmented non-body components to create a personalized complete model SMPLX++.
\begin{figure}[ht]
    \centering
    \includegraphics[width=1.0\linewidth]{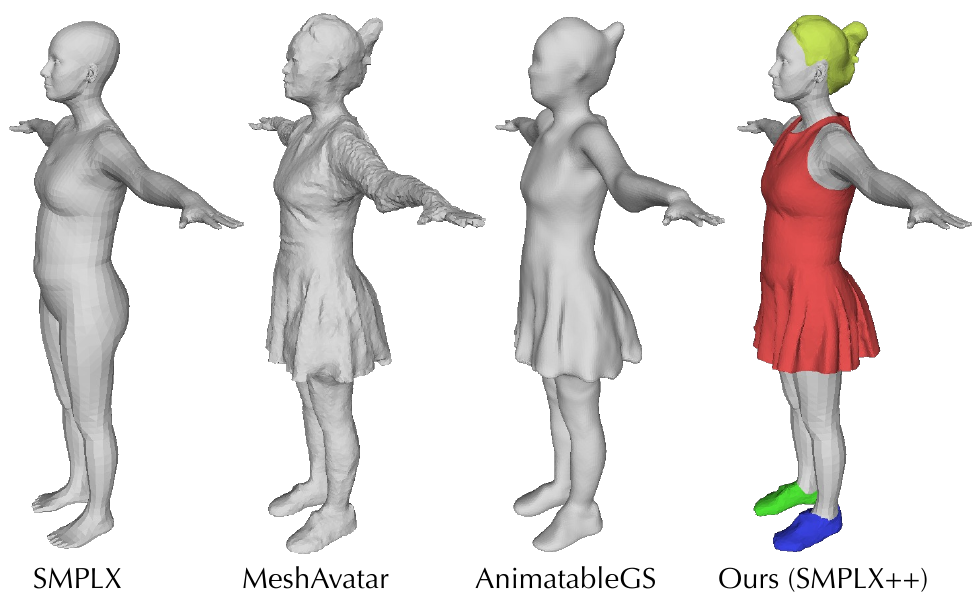}
    \caption{\textbf{Template Comparison.}}
    \label{fig:template_comparison}
\end{figure}
\begin{figure}[ht]
    \centering
    \includegraphics[width=1.0\linewidth]{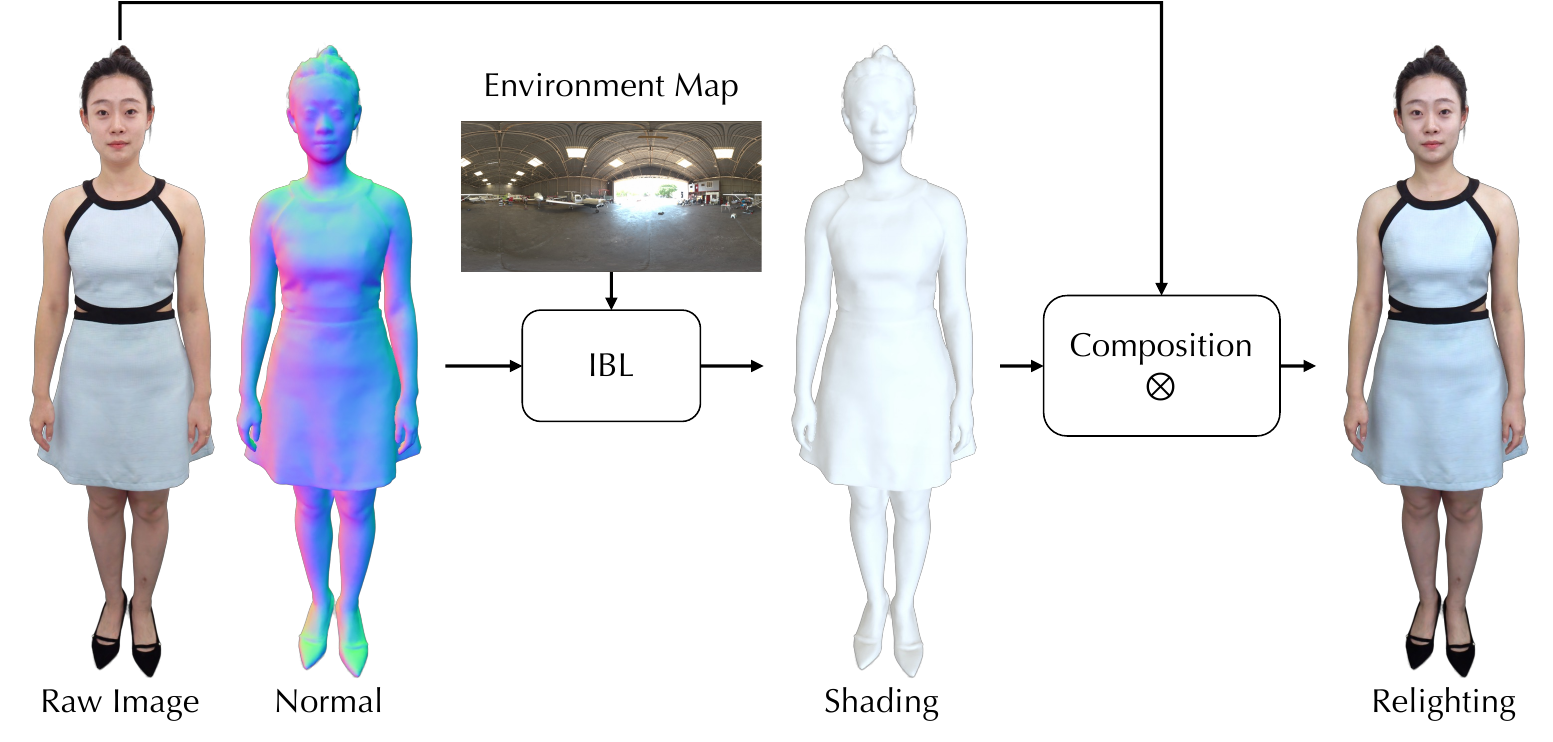}
    \vspace{-1.0em}
    \caption{\textbf{Relighting Visualisation.}}
    \label{fig:relighting}
\end{figure}
The parametric template SMPLX++ can be driven by expression and pose parameters same as the naive SMPLX, which is more expressive for loose clothing geometry.
The template contains roughly $23k$ vertices and $45k$ faces, including $20k$ faces for clothes, $2k$ faces for hair, and $2k$ faces for shoes.
In contrast to MeshAvatar~\cite{meshavatar} and AnimatableGS~\cite{animatable_gaussian}, which learn an implicit template from scratch, our template preserves a priori facial expressions and hand gestures as shown in~\cref{fig:template_comparison}, which are essential for achieving natural and expressive animations.

\begin{figure*}[ht]
    \centering
    \includegraphics[width=0.75\linewidth]{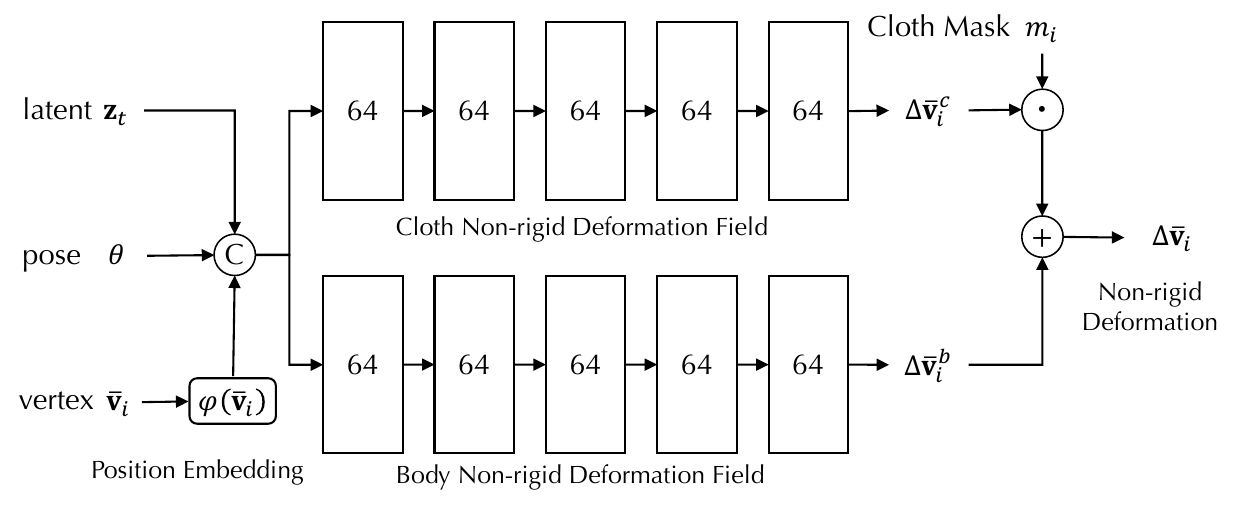}
    \caption{\textbf{Network Architecture of Mesh Nonrigid Deformation Field.}}
    \label{fig:nonrigid_mlp}
\end{figure*}
\begin{table*}[ht]
	\centering
	\scalebox{0.9}{
		\begin{tabular}{c|c|c|c|c|c|c|c|c|c}
			\toprule
            
			 \multirow{2}{*}{Model} & \multirow{2}{*}{Template} & \multirow{2}{*}{Non-rigid} & \multicolumn{2}{c|}{Gaussian/Face Num.} & \multicolumn{2}{c|}{Quality} & \multicolumn{2}{c|}{Controllability} & Speed \\
             \cline{4-9}
            & & & Head & Body & Head & Body & Head & Body & (Inference)\\
            \hline
            3DGS-Avatar~\cite{3dgs-avatar} & SMPLX & MLP & 19k & 181k & low & low & low & low & 54 \\
            GaussianAvatar~\cite{gaussianavatar} & SMPLX  & Unet & 45k & 146k & low & low & low & medium & 55 \\
            MeshAvatar~\cite{meshavatar} & Mesh (Implicit) & StyleUnet & 5k & 50k & low & medium & low & medium & 22 \\
		AnimatableGS~\cite{animatable_gaussian} & Mesh (Implicit) & StyleUnet & 18k & 246k & low & high & low & high & 16 \\
            \hline
            Ours (Teacher)  & SMPLX++ & StyleUnet & 19k & 250k & medium & high & medium & high & 16 \\
            Ours (Student)  & SMPLX++ & MLP+BS & 70k & 200k & high & high & high & high & 156 \\
			
			\bottomrule
	\end{tabular}}
	\caption{{\bf Summary about these State-of-the-art Methods of Full-body Avatars}.}
	\label{tab:summary}
\end{table*}

\noindent \textbf{Network Architecture.}
We employ a compact MLP-based student network to learn the pose-dependent non-rigid deformation of mesh:
\begin{equation}
\begin{aligned}
    \mathbf{g}_{i} &= \varphi\left(\bar{\mathbf{v}}_{i}\right)\oplus \mathbf{\theta} \oplus \mathbf{z}_{t} \\
    \Delta\bar{\mathbf{v}}_{i} &= \mathcal{S}_{c}\left(\mathbf{g}_{i}\right)\cdot m_i + \mathcal{S}_{b}\left(\mathbf{g}_{i}\right)
\end{aligned}
\end{equation}
where $\bar{\mathbf{v}}_{i} \in \mathbb{R}^3$ is the $i$-th vertex coordinate in the canonical space, $\theta \in \mathbb{R}^{63}$ is the pose parameter, and $\mathbf{z}_t \in \mathbb{R}^{32}$ is a learnable embedding for each frame to compensate for inaccurate pose estimation.
The positional encoding function $\varphi\left(\cdot\right)$  introduced in NeRF~\cite{nerf}, is applied with a frequency band of $L=6$ in our experiments.
The architecture of the student network comprises two specialized MLPs.
The first MLP, $\mathcal{S}_{b}$, models the body's non-rigid deformations, while the second MLP, $\mathcal{S}_{c}$, captures additional deformations arising from clothing dynamics.
To ensure that clothing deformations are applied exclusively to vertices associated with clothing, we introduce a mask $m_i \in \{0, 1\}$, where $m_i=1$ for the vertices belonging to clothing.

\noindent \textbf{Relighting.} We ensure that ambient lighting around the performer is as uniform and white as possible during capture. We use the raw rendered image as the base color and apply shading with new environment light based on the rendered normal map as shown in~\cref{fig:relighting}. Although this approach is not physically accurate, it results in better integration with the environment.

\begin{figure*}[ht]
    \centering
    \includegraphics[width=1.0\linewidth]{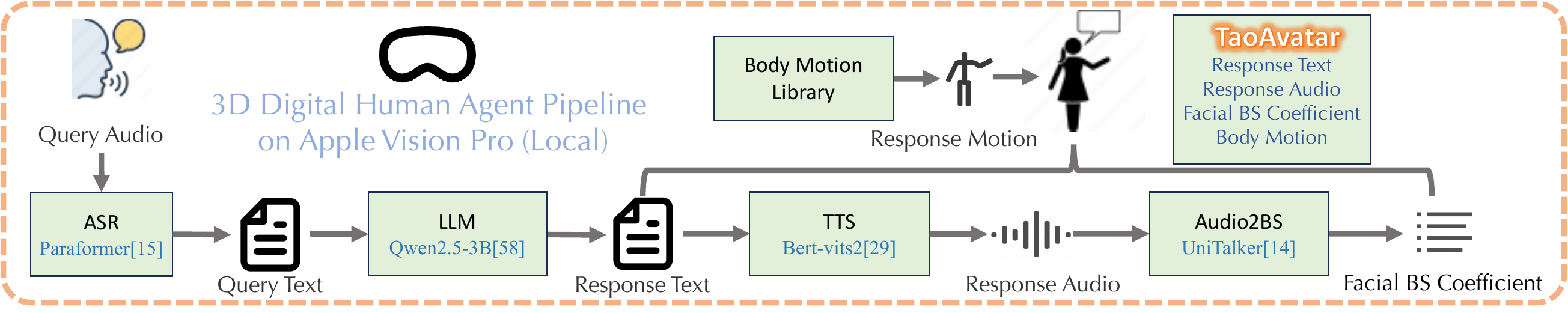}
    \caption{\textbf{3D Digital Human Agent Pipeline.}}
    \label{fig:AIAgent_Pipeline}
\end{figure*}

\noindent \textbf{Deployment and 3D Digital Human Agent Pipeline.} We make some efforts for mobile deployment, primarily including: a) FP16 quantization for the MLP; b) UInt16 quantization for Gaussian sorting; and c) asynchronous inference techniques, where the animation system operates at 20 FPS (capture frame rate of training data) while the rendering system interpolates animations to render at 90 FPS (maximum screen refresh rate) on the Apple Vision Pro. Please note that all these strategies are not applied on RTX4090 in~\cref{tab:novel_view_and_novel_pose_on_talking_body}, which can fundamentally demonstrate the performance of our method. We develop a 3D digital human agent on the Apple Vision Pro, which interacts with users through an ASR-LLM-TTS-Audio2BS pipeline~\cite{Paraformer,Qwen2,VITS2,unitalker} as shown in ~\cref{fig:AIAgent_Pipeline}. Notably, all models run locally after deployment. Please stay tuned for future work with more technical details.

\begin{figure}[ht]
    \centering
    \includegraphics[width=1.0\linewidth]{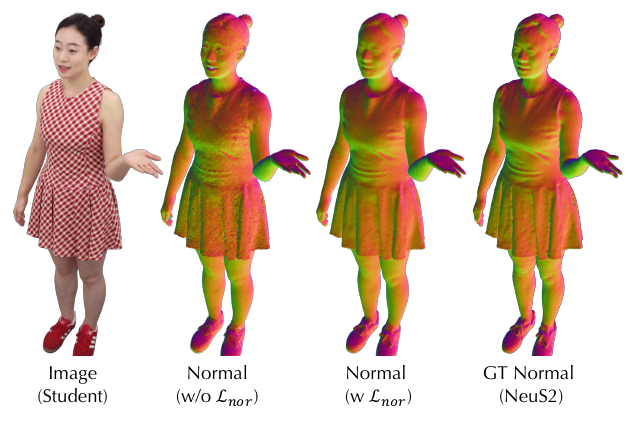}
    \vspace{-2.0em}
    \caption{\textbf{The Impact of Normal Loss.}}
    \label{fig:normal_loss}
\end{figure}

\begin{figure*}[ht]
    \centering
    \includegraphics[width=1.0\linewidth]{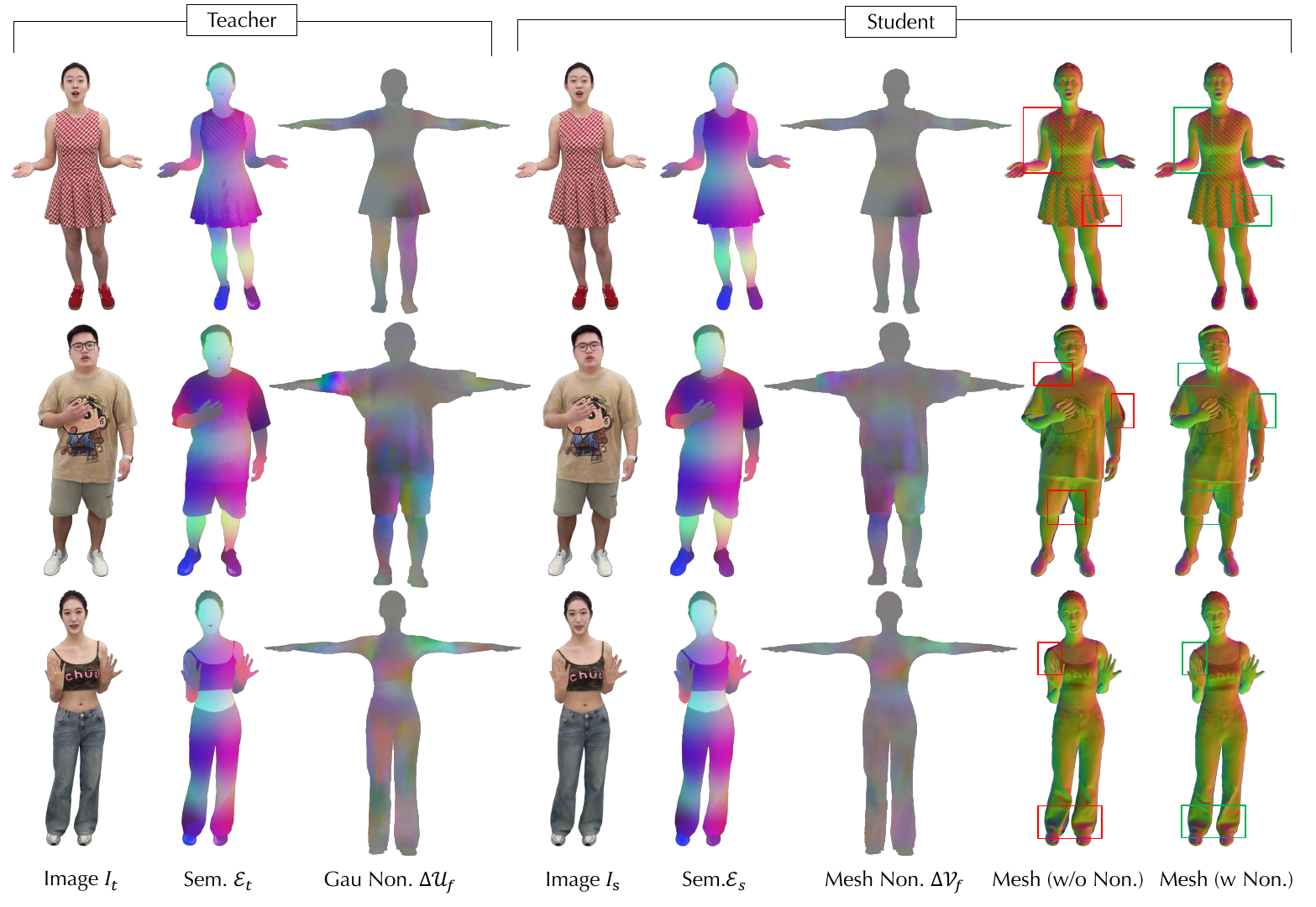}
    \vspace{-2.0em}
    \caption{\textbf{Qualitative Visualization of Baking.}}
    \label{fig:baking_results}
\end{figure*}

\noindent \textbf{Reanimation.} We introduce a learnable embedding $\mathbf{z}_t$ for each frame to better compensate for misalignment issues caused by the inaccurate SMPLX~\cite{smplx} estimation and dynamic changes that cannot be captured by body pose $\mathbf{\theta}$ (e.g., clothing inertia and swing, changes in hand muscles, etc.). For offline reanimation, we can utilize the Nonrigid Deformation Baking method introduced in the paper to obtain the corresponding $\mathbf{z}_t$ under novel poses from the teacher network. For online real-time body driving, we use the $\mathbf{z}_0$ from the first training frame, which is practically acceptable, although it compromises the accuracy of non-rigid deformations.

\section{Experiment Details}

\begin{figure*}[ht]
    \centering
    \includegraphics[width=1.0\linewidth]{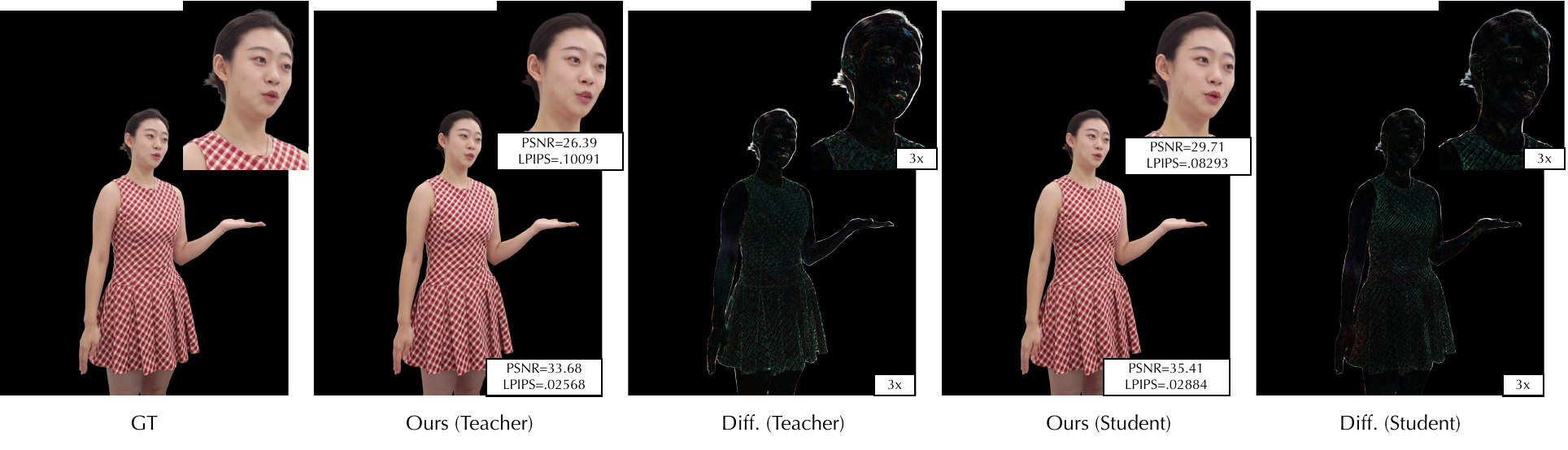}
    \caption{\textbf{Qualitative Comparison of Details.}}
    \label{fig:metric_comparison}
\end{figure*}

\noindent \textbf{Metric Evaluation.} To quantitatively evaluate the quality of the rendered images, we choose Peak Signal-to-Noise Ratio (PSNR), Structure Similarity Index Measure (SSIM)~\cite{SSIM}, and Learned Perceptual Image Patch Similarity (LPIPS)~\cite{lpips}.
In our experiments, we evaluate masked images at a resolution of $1500\times 2000$, where the masks are provided by BiRefNet~\cite{birefnet}.
It is important to note that while PSNR and SSIM are highly sensitive to pixel misalignment, LPIPS demonstrates greater robustness by computing differences in deep feature maps.
As illustrated in~\cref{fig:metric_comparison}, the teacher network delivers superior full-body clothing details (better LPIPS scores), while the student network excels in the face region due to a more plausible Gaussian distribution.
This discrepancy primarily arises from background residuals introduced by the segmentation network~\cite{birefnet} and the teacher network's propensity to omit fine details, such as fragmented hair strands.
Additionally, we crop the face region for evaluation based on the projection of the head bounding box. 
We also adopt point-to-surface distance (P2S) and Chamfer distance (Chamfer) to evaluate the geometry, while the ground truth mesh is generated from NeuS2~\cite{neus2}.

\begin{figure}[ht]
    \centering
    \includegraphics[width=1.0\linewidth]{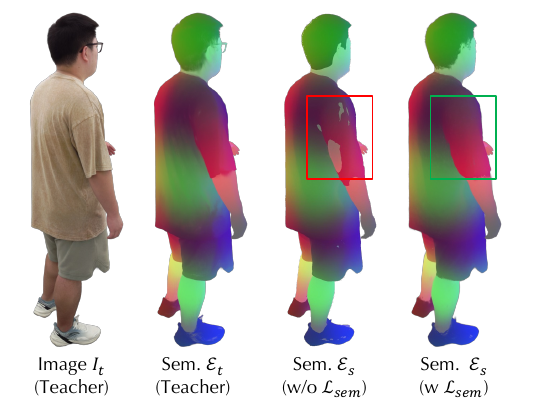}
    \vspace{-2.0em}
    \caption{\textbf{Ablation Study on Semantic Loss during Baking.}}
    \vspace{-1.0em}
    \label{fig:semantic_loss}
\end{figure}

\section{Additional Discussion}

\noindent \textbf{Discussion w.r.t AnimatableGS.}
In our teacher-student framework, we utilize AnimatableGS~\cite{animatable_gaussian} as the teacher network due to its robust capability to model complex pose-dependent non-rigid deformations. Unlike the original AnimatableGS~\cite{animatable_gaussian}, which learns an implicit template from scratch, we adopt the SMPLX++ model as our predefined template. We input semantic positional maps into StyleUnet~\cite{styleavatar} by assigning distinct color labels to each component (e.g., red for clothing, yellow for hair) and combining these labels with the posed coordinates to generate the final vertex colors.
This strategy can provide better semantic information about segmentation for the teacher network.
Additionally, we incorporate a normal loss $\mathcal{L}_{nor}$ during the training of the teacher network, which contributes to the learning of smoother geometries.

\begin{figure*}[ht]
    \centering
    \includegraphics[width=1.0\linewidth]{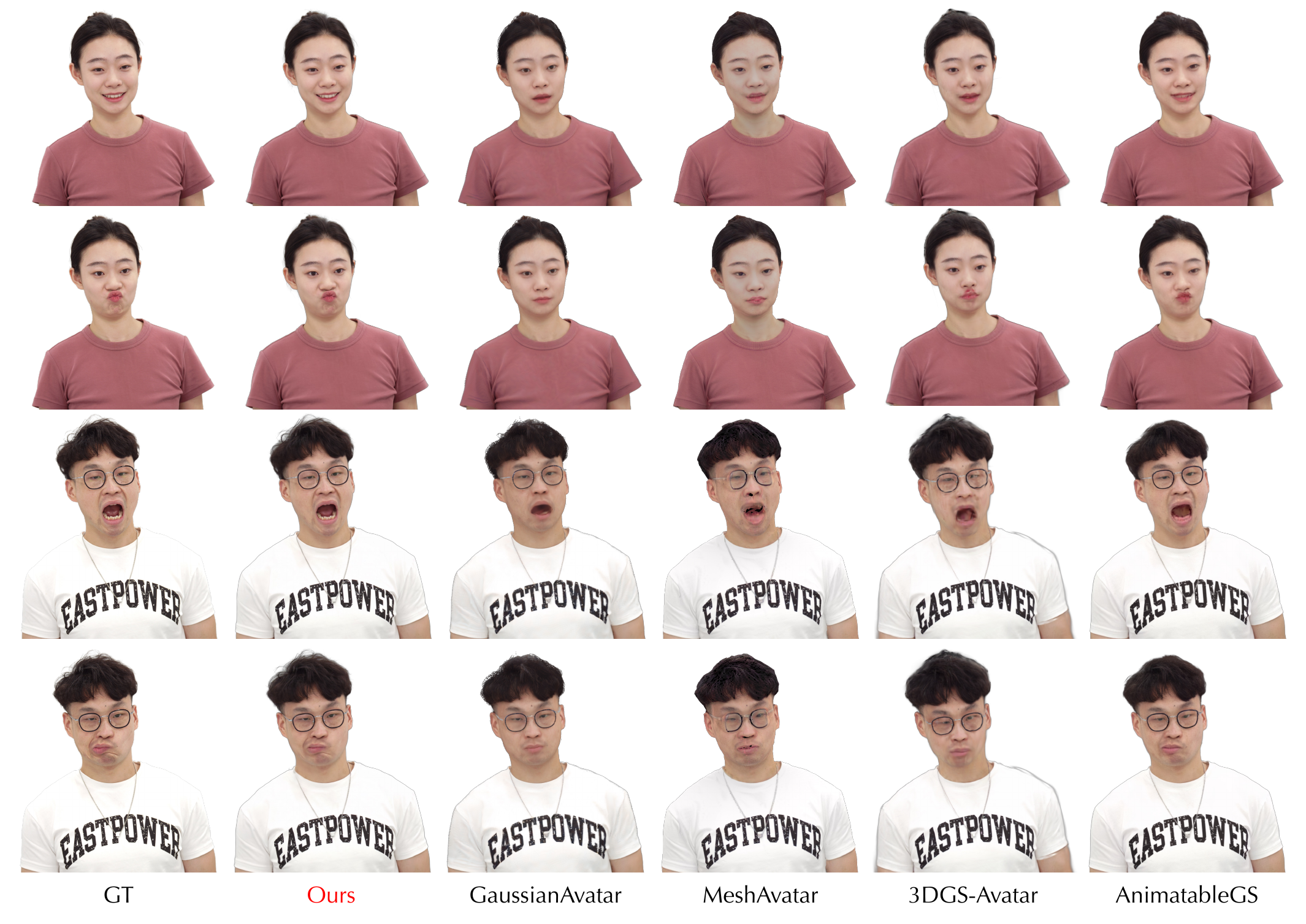}
    \caption{\textbf{Qualitative Comparisons on Exaggerated Expression.}}
    \label{fig:expression_comparison}
\end{figure*}

\noindent \textbf{Summary about Full-body Avatars.}
We present a comparative summary of full-body avatar methods in~\cref{tab:summary}. 3DGS-Avatar~\cite{3dgs-avatar} and GaussianAvatar~\cite{gaussianavatar} utilize the basic naked SMPLX model as their template, resulting in poor rendering quality for loose clothing.
MeshAvatar~\cite{meshavatar} and AnimatableGS~\cite{animatable_gaussian} develop implicit clothed templates from scratch which compromising the control over facial expressions and hand gestures. Regarding non-rigid deformation modeling, StyleUnet exhibits more robust expressive capabilities than MLP and Unet, as discussed in AnimatableGS~\cite{animatable_gaussian}. Our method employs an MLP-based student network baked from the teacher network and two lightweight learnable blend shape compensations.
This design enables high-performance rendering with minimal quality degradation.
Notably, while maintaining the same number of Gaussians as the teacher model, we allocate more Gaussians to the face to achieve higher facial sharpness as shown in~\cref{fig:expression_comparison}.
In contrast, AnimatableGS~\cite{animatable_gaussian} limits the number of Gaussians on the face due to the resolution constraints of the rendered positional maps.

\noindent \textbf{The Non-rigid Loss and Semantic Loss during Baking.}
During the non-rigid deformation baking process, both the non-rigid loss $\mathcal{L}_{non}$ and the semantic loss $\mathcal{L}_{sem}$ play crucial roles.
For the non-rigid deformation loss $\mathcal{L}_{non}$, we directly use the Gaussian non-rigid deformation maps $\{\Delta\mathcal{U}_{f},\Delta\mathcal{U}_{b}\}$ generated by the teacher network to directly supervise the mesh non-rigid deformation maps $\{\Delta\mathcal{V}_{f},\Delta\mathcal{V}_{b}\}$ of the student network under T-pose in the canonical space.
Regarding the semantic loss $\mathcal{L}_{sem}$, we construct a semantic label $\mathbf{e}_{i} = \mathbf{c}_{i} + \sin{\left(\tau\bar{\mathbf{v}}_i\right)}$ for each vertex of the template, where $\tau$ is a scale factor is designed to increase the frequency of positional changes, inspired by the position embedding in NeRF~\cite{nerf}.
As illustrated in~\cref{fig:semantic_loss}, the semantic loss $\mathcal{L}_{sem}$ helps mitigate the intersection between clothing and the body.
We provide visualizations of the products from the baking process of different identities in~\cref{fig:baking_results}.
The teacher network effectively guides learning mesh non-rigid deformation of the student network, resulting in geometry that is well-aligned with the performer's surface as shown in~\cref{fig:baking_results} (Mesh (w/o Non.) vs. Mesh (w Non.)).
Without the help of the baking process, it isn't easy to decouple geometry and appearance.

\noindent \textbf{The Impact of Normal Loss.}
Similar to most 3D Gaussian-based methods~\cite{gaussianshader,gs_ir}, we define the direction of the Gaussian's shortest axis as its normal.
In contrast to other approaches that dynamically determine the normal orientation based on the camera position, we assign a fixed normal $n=[1,0,0]$ and scaling $s=[\epsilon,1,1]$ in the local space for each Gaussian, where $\epsilon=0.01$ is a minimum to make the Gaussian as thin as possible.
Upon transforming its parent triangle, the Gaussian's normal in world space aligns with the triangle's normal.
To promote smoother rendered normal maps,  we introduce a normal loss $\mathcal{L}_{nor}$ as illustrated in~\cref{fig:normal_loss}.
The ground truth normal maps are obtained from NeuS2~\cite{neus2}.
Additionally, the rendered normal maps facilitate image-based relighting, as demonstrated in the provided video demo.


\section{Failure Cases}
\begin{figure}[ht]
    \centering
    \includegraphics[width=1.0\linewidth]{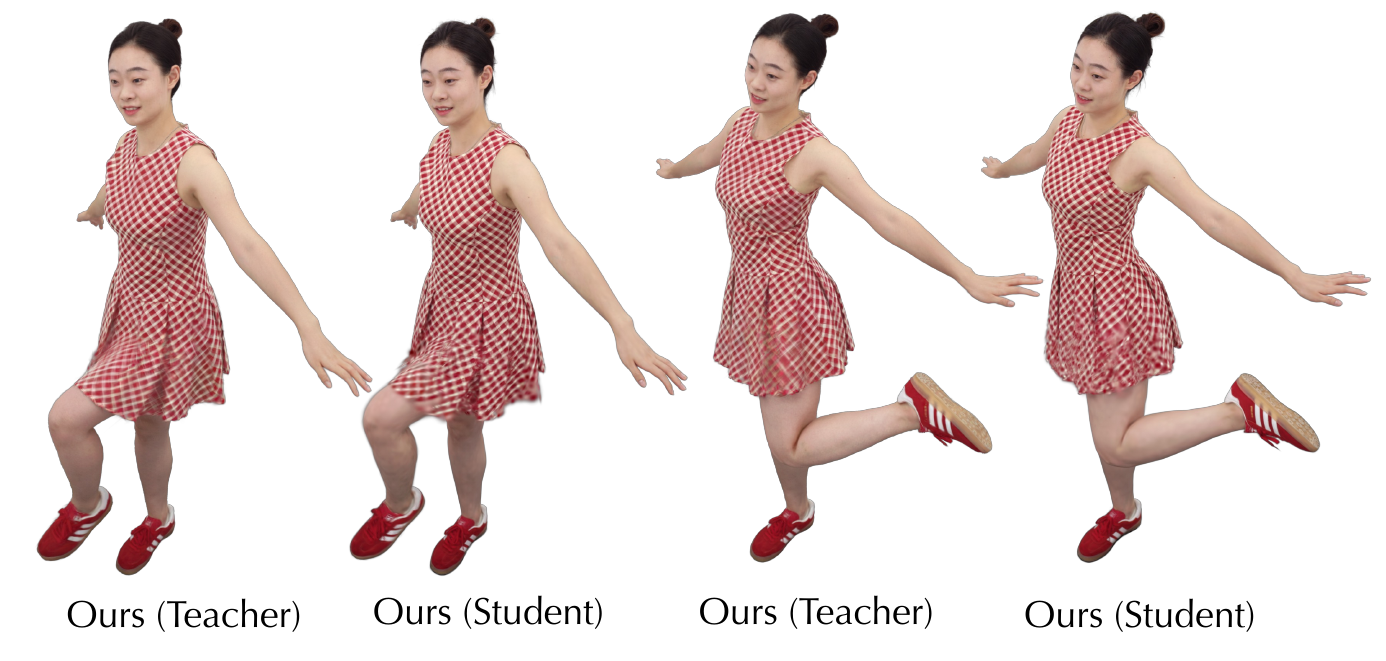}
    \vspace{-2.0em}
    \caption{\textbf{Failure Cases.}}
    \vspace{-1.0em}
    \label{fig:failure_cases}
\end{figure}

Although TaoAvatar demonstrates remarkable performance in full-body talking tasks, it still faces challenges in handling complex motions and exaggerated outfits.
Specifically, when the teacher network struggles to accurately model loose garments under intricate motions (e.g., dancing-induced skirt motions), the task becomes increasingly difficult for the student network as shown in~\cref{fig:failure_cases}.
Moreover, TaoAvatar is highly reliant on the precision of SMPLX parameters and is susceptible to artifacts when the estimated SMPLX fails to align with the image properly.

\clearpage

\end{appendix}

\end{document}